# Using Dempster-Shafer Theory in Knowledge Representation

Alessandro Saffiotti*

IRIDIA - Université Libre de Bruxelles
Av. F. Roosevelt 50 - CP 194/6
B-1050 Bruxelles - Belgium
E-mail: r01507@bbrbfu01.bitnet

## Abstract

In this paper, we suggest marrying Dempster-Shafer (DS) theory with Knowledge Representation (KR). Born out of this marriage is the definition of "Dempster-Shafer Belief Bases", abstract data types representing uncertain knowledge that use DS theory for representing strength of belief about our knowledge, and the linguistic structures of an arbitrary KR system for representing the knowledge itself. A formal result guarantees that both the properties of the given KR system and of DS theory are preserved. The general model is exemplified by defining DS Belief Bases where First Order Logic and (an extension of) KRYPTON are used as KR systems. The implementation problem is also touched upon.

## 1. Introduction

Dempster-Shafer (hereafter, DS) theory of evidence is a mathematical setting for expressing the strength of our belief over a set of possible hypotheses (Shafer, 1976; Smets, 1988). In a nutshell, a unitary amount of belief ("belief mass") is distributed among the subsets of a set $\Theta$ of (exhaustive and mutually exclusive) hypotheses, called the *frame of discernment*. A *Basic Probability Assignment* (BPA) over $\Theta$ is a function specifying how heavy a mass of belief is being allocated to each of its subsets. A *belief function* $bel_m$ is then defined on the subsets of $\Theta$: intuitively, $bel_m(A)$ measures the maximum amount of belief that can be allocated to A according to $m$. Two BPA's on the same space of hypotheses, corresponding to two distinct bodies of evidence, may be combined into a new BPA by means of *Dempster's rule of combination*. Acquisition of new evidence is typically performed by combining a pre-existing BPA with a new BPA representing the impact of the new evidence alone. A most popular use of DS theory in AI is well exemplified by what has been called the "multivariate formalism" (Kong, 1986): the problem we want to deal with is represented through a number of variables, associated to the relevant elements of the problem. Each variable is allowed to take a number of alternative values, and BPA's are defined over the sets of these values. A relation among different elements is expressed by defining a BPA on the product space of the (values of the) corresponding variables. Once we have modelled a problem this way, a DS solution is found by computing an overall BPA—the combination of all the available BPA's (i.e. those expressing prior belief, those expressing relations, and those expressing acquired evidence)—on the overall space of values for all the variables. From this BPA, a measure of belief for the (sets of) values of each interesting variable can be computed.

Expressive power and epistemic adequacy have often been acknowledged to DS theory in AI. Unfortunately, uncertainty representation is but one side of the problem of formalizing uncertain knowledge: in particular, the knowledge itself, which uncertainty refers to, has to be represented. If we take the point of view of Knowledge Representation (KR), we may drive the conclusion that DS theory is not an adequate tool for expressing general or domain specific <u>knowledge</u>. Different types of knowledge have continuously been identified in the KR tradition (see e.g. Israel & Brachman, 1981; Brachman & Levesque, 1982); to wit, the four statements

- Elephants are mammals with four legs
- All of my friends like jazz music
- Normally, birds fly
- Smoke suggests fire

express four qualitatively different types of knowledge. A number of efforts have normally been devoted in KR to the development of representation formalisms able to capture (and discriminate) different types of knowledge. Yet, in a DS framework, all the four statements above would most probably be expressed by a single pattern, namely a BPA on the product space of two appropriate variables (or a conditional belief function $bel(A|B)$ over it). Moreover, representing even simple patterns of generic knowledge in a DS framework may become highly problematic: e.g., consider representing the fact that parents are humans that have at least one child (notice that this fact would be easily represented in most KR system).

In this paper we try to turn DS theory to KR. We do this by defining a formal framework which uses DS theory for representing strength of belief about our knowledge, and the linguistic structures of an arbitrary KR language for representing the knowledge itself. The key concept consists in the distinction between the language we use for representing knowledge, peculiar of the given KR system, and the knowledge actually represented. This knowledge will be modelled by abstract objects (*propositions*), and belief about it will be ex-

* This research has been partially supported by the ARCHON project, funded by grants from the Commission of the European Communities under the ESPRIT-II Program, P-2256.

pressed as BPA's on these objects (rather than on the sentences of the KR language). As a consequence, our framework will be independent of the particular KR system we intend to use, regarded as a parameter. In order to settle down our framework in a general way, we will adhere to the "functional approach" advocated by Levesque (1984), and define "Dempster-Shafer Belief Bases" as abstract data types representing a (uncertainly) believed corpus of knowledge. In them, knowledge and belief about this knowledge will be represented according to a specific KR language and to DS theory, respectively.

The rest paper is organized as follows. Propositions, and the way we use them to bridge DS theory and KR systems, are presented in Section 2. Section 3 gives the formal account of Belief Bases, together with some results showing their "good behaviour". Section 4 shows two possible ways (model-theoretic and proof-theoretic) of modelling Belief Bases; also, it illustrates their use and behaviour by giving some examples. Section 5 deals with the possibility to turn the proposed framework into a (implementable) hybrid system. Section 6 discusses some related work, and concludes.

## 2. The Objects of Belief

In introducing our framework in the above section, the word "knowledge" has been used in two ways: to denote the objects which we entertain belief in; and to denote the objects represented by a KR language. We analyse in this section how these "objects" can be modelled in an abstract way, and how they can become the link between DS theory and a KR system.

We identify items of knowledge with elements of a set $\mathcal{P}$ of "propositions"[1]. In order for $\mathcal{P}$ to be a good candidate for our goal, we actually require it to be a Boolean algebra; we denote by $\Rightarrow$ its partial order, by $\cap$ its infimum, by $^c$ its complement, and by $\mathcal{U}$ and $\emptyset$ its top and bottom elements, respectively. Given elements P and Q of $\mathcal{P}$, we read "P $\Rightarrow$ Q" as P *logically entails* Q, "P $\cap$ Q" as the *conjunction* of P and Q, $^cP$ as the *negation* of P, $\mathcal{U}$ as the *tautology* and $\emptyset$ as the *false proposition* of $\mathcal{P}$. Propositions are the right objects to asses belief to, so we will define a DS calculus on $\mathcal{P}$ in the next section. However, we want to express our knowledge through the sentences of a generic KR language: the problem we face is then how to map sentences of this language to the propositions they connote. We will suggest that providing this mapping actually constitutes the role of the KR system in the framework that we are defining; but we first need to spend a few words on KR systems and their semantics.

The most popular way to formally describe a KR system $\Sigma$ is to define a formal system based on the language $\mathcal{L}_\Sigma$ of $\Sigma$. There are basically two approaches to formalizing such a system, namely the model-theoretic and the proof-theoretic ones. The first approach consists in providing a truth relation $\vDash_\Sigma$ for $\Sigma$, which defines the notion of "truth" of sentences of $\mathcal{L}_\Sigma$ with respect to elements of a given class of mathematical structures, used as models of the language. The second approach consists in providing $\Sigma$ with a deduction relation $\vdash_\Sigma$ that specifies which sentences can be deduced by which ones through the deductive apparatus of $\Sigma$ (normally a set of axioms and inference rules). Correspondingly, we indicate a (formal account of a) KR system by $\Sigma = (\mathcal{L}_\Sigma, \vDash_\Sigma)$ or by $\Sigma = (\mathcal{L}_\Sigma, \vdash_\Sigma)$. For the goals of this paper, we regard the role of a KR system $\Sigma$ as that of 1) providing a KR language $\mathcal{L}_\Sigma$, and 2) mapping each sentence of $\mathcal{L}_\Sigma$ to the proposition it connotes (i.e. attributing a meaning to it). So, given an algebra of propositions $\mathcal{P}$ and a KR system $\Sigma$, we are interested in a function

$$\|\cdot\|_\Sigma : \mathcal{L}_\Sigma \to \mathcal{P}$$

which maps each sentence $\alpha$ of $\mathcal{L}_\Sigma$ to the proposition $\|\alpha\|_\Sigma \in \mathcal{P}$ connoted by $\alpha$ according to $\Sigma$. We call $\|\cdot\|_\Sigma$ a *meaning function* for $\Sigma$. We will see in Section 4 how we can define this function for a given formal account of a KR system.

DS Belief Bases, to be defined shortly, will use the mapping $\|\cdot\|_\Sigma$ to bridge $\mathcal{L}_\Sigma$ and $\mathcal{P}$. As a matter of fact, belief will be entertained—and manipulated through DS theory—on the space $\mathcal{P}$ of propositions, but we will use the language $\mathcal{L}_\Sigma$ to express our knowledge. The advantage of this is that the whole framework may be defined in general terms, irrespective of the particular $\mathcal{L}_\Sigma$ that we decide to use. The situation is illustrated below.

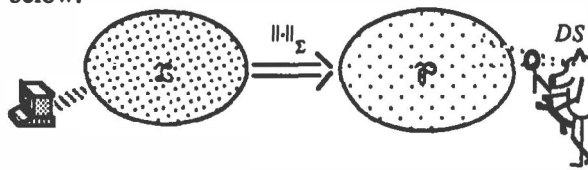

## 3. Dempster-Shafer's Belief Bases

In proposing his "functional approach to Knowledge Representation", Levesque (1984) suggested modelling knowledge by means of abstract data types ("Knowledge Bases") characterized in terms of the op-

---

[1] We remind that, after Frege, the *proposition* connoted by a sentence is its *sense*, its *meaning*. The reader should never confuse propositions with sentences which express them; nor should she confuse them with propositional symbols, or any other syntactical element of any given language. Propositions are the <u>content</u> of the sentences.

erations that can be made on them; typical operations are a query operation "Ask", an update operation "Tell", and an initialization operation "Empty". Adapting Levesque's suggestion to the case of uncertain knowledge, we will define "Dempster-Shafer's Belief Bases" (or "Belief Bases") as abstract data types where knowledge and belief about this knowledge are represented according to a specific KR language and to DS theory, respectively. Belief Bases for a given KR system $\Sigma$ will be characterized by three primitive operations ($\mathcal{BB}$ denotes the class of Belief Bases):

$\text{Ask}_\Sigma : \mathcal{L}_\Sigma \times \mathcal{BB} \to ([0,1] \times [0,1])$

$\text{Tell}_\Sigma : \mathcal{L}_\Sigma \times ([0,1] \times [0,1]) \times \mathcal{BB} \to \mathcal{BB}$

$\text{Empty}_\Sigma : \{\emptyset\} \to \mathcal{BB}$

whose intended semantic is as follows. $\text{Ask}_\Sigma$ determines to what extent a sentence is believed (according to DS theory) to hold (according to the logic of $\Sigma$) in a Belief Base; $\text{Tell}_\Sigma$ returns a new Belief Base obtained by acquiring a new piece of evidence; and $\text{Empty}_\Sigma$ provides an "empty" Belief Base, where nothing is believed except the valid sentences of $\mathcal{L}_\Sigma$.

In order to define the above operations more precisely, we have to specify the structure of Belief Bases. Given a Boolean algebra $\mathcal{P}$ of propositions, we will model Belief Bases as sets of pairs $<P, m_P>$ with $P$ an element of $\mathcal{P}$ and $m_P$ a real number in $[0,1]$. Intuitively, each Belief Base represents a BPA on $\mathcal{P}$ [2]. Correspondingly, $\text{Empty}_\Sigma[]$ will return a vacuous BPA, allocating all the mass to the tautology $\mathcal{U}$; the pair returned by $\text{Ask}_\Sigma[\alpha,\kappa]$ will contain the values of $bel_\kappa$ for the asked proposition[3] and its complement, respectively; and $\text{Tell}_\Sigma[\alpha,<x_t,x_f>,\kappa]$ will return the new BPA obtained by combining, by Dempster's combination, the old BPA $\kappa$ with a BPA representing a piece of evidence saying: «I believe to a degree $x_t$ that $\alpha$ is true, and to a degree $x_f$ that it is false". Before going to the formal definition, we need to restate some of the basic elements of DS theory in terms of our generic Boolean algebra $\mathcal{P}$ (with $\Rightarrow, \cap, ^c, \mathcal{U}$ and $\emptyset$ as above) and of sets of pairs $<P, m_P>$[4]:

---

[2] As reminded in the Introduction, the usual formulation of DS theory is given starting with a set $\Theta$ of hypotheses, and by then defining BPA's and belief functions over the power-set of $\Theta$ (which is a Boolean algebra). Here, on the contrary, we formulate the theory directly on a generic Boolean algebra $\mathcal{P}$. The two formulations are equivalent.

[3] Short for "the proposition connoted by the asked sentence".

[4] $\text{Bel}_\kappa(\mathcal{U}) = 1$ and $\text{Bel}_\kappa(\emptyset) = 0$ will follow from def. 3.3 below.

**Def. 3.1.** *Let $\mathcal{P}$ be a Boolean algebra as above, and let $\kappa$ be a set of pairs $<P, m_P>$, with $P \in \mathcal{P}$ and $m_P \in [0,1]$. Then, for any $Q \in \mathcal{P}$, we let*

$\text{Bel}_\kappa(Q) =_{df} \sum \{m_P \mid <P, m_P> \in \kappa \text{ and } P \Rightarrow Q\}$

**Def. 3.2.** *Let $\mathcal{P}$ be a Boolean algebra and $\kappa_1$ and $\kappa_2$ two sets of pairs as above. Then, <u>Dempster's combination</u> of $\kappa_1$ and $\kappa_2$ is given by*

$$\kappa_1 \oplus \kappa_2 =_{df} \left\{ <Q, m_Q> \mid m_Q = \rho_{12}^{-1} \sum_{\substack{Q = P' \cap P'' \\ <P', m_{P'}> \in \kappa_1, <P'', m_{P''}> \in \kappa_2}} m_{P'} m_{P''} \right\}$$

*if $\rho_{12} \neq 0$, it is undefined otherwise; $\rho_{12}$ is given by:*

$\rho_{12} = 1 - \sum m_{P'} m_{P''}$ *for all* $<P', m_{P'}> \in \kappa_1, <P'', m_{P''}> \in \kappa_2$ *s.t.* $P' \cap P'' = \emptyset$

We have now all the ingredients for giving the functional definition of Belief Bases for a given KR system.

**Def. 3.3.** *Let $\mathcal{P}$ be a Boolean algebra of propositions as above. Let $\Sigma$ be a KR system and $\|\cdot\|_\Sigma$ a meaning function for $\Sigma$. Then <u>Dempster-Shafer Belief Bases</u> on $\Sigma$ are defined by the following operations:*

$\text{Empty}_\Sigma[] =_{df} \{<\mathcal{U}, 1>\}$

$\text{Ask}_\Sigma[\alpha, \kappa] =_{df} < \text{Bel}_\kappa(\|\alpha\|_\Sigma), \text{Bel}_\kappa(^c\|\alpha\|_\Sigma) >$

$\text{Tell}_\Sigma[\alpha, <x_t,x_f>, \kappa] =_{df}$

$= \kappa \oplus \{<\|\alpha\|_\Sigma, x_t>, <^c\|\alpha\|_\Sigma, x_f>, <\mathcal{U}, 1-x_t-x_f>\}$

*provided that, in $\text{Tell}_\Sigma$, the evidence represented by $\alpha$ is distinct from that represented by $\kappa$.*

A Belief Base on $\Sigma$ is built starting with $\text{Empty}_\Sigma[]$, and then by performing successive $\text{Tell}_\Sigma$ operations on it. Hence, as desired, a Belief Bases on $\Sigma$ turns out to be a set of a pairs $<P, m_P>$, with $P \in \mathcal{P}$ and $m_P \in [0,1]$, such that all the $m_P$'s add up to 1. We interact with Belief Bases, via the above operations, by using <u>sentences</u> of $\mathcal{L}_\Sigma$ for expressing our knowledge, and numbers in $[0,1]$ for expressing our belief in the truth (or falsity) of this knowledge. However, we emphasise once again that mass of belief is actually allocated to the <u>propositions</u> connoted by these sentences, and not to the <u>sentences</u> themselves. We will see examples of Belief Bases in the next sections. Nevertheless, just to keep the curiosity of the reader burning, we show here a typical use of a Belief Base where First Order Logic has been chosen to represent knowledge.

$\text{Ask}_{\text{FOL}}[Q(x), \text{Tell}_{\text{FOL}}[P(x), <0.6, 0.3>,$

$\quad \text{Tell}_{\text{FOL}}[\forall x. P(x) \supset Q(x), <0.8, 0>, \text{Empty}_{\text{FOL}}[]]]] =$

$= <0.48, 0>$

Belief Bases "behave well" as hybrid structures; that is, they preserve all the properties of both DS theory and the used KR system. For instance, we have[5]:

**Lemma 3.1.**

i) $\text{Ask}_\Sigma[\alpha, \text{Empty}_\Sigma[]] = \begin{cases} <1,0> & \text{if } \alpha \text{ is valid in } \Sigma; \\ <0,0> & \text{otherwise} \end{cases}$

ii) $\text{Ask}_\Sigma[\alpha, \text{Tell}_\Sigma[\beta, <1,0>, \text{Empty}_\Sigma[]]] =$
$= \begin{cases} <1,0> & \text{if } \beta \text{ logically implies } \alpha \text{ in } \Sigma; \\ <0,0> & \text{otherwise} \end{cases}$

So, Belief Bases on $\Sigma$ obey the logic of $\Sigma$, and they just mirror this logic when belief measures are restricted to be 0 or 1. On the other hand, any DS model can be captured in our framework:

**Theorem 3.1.** *Let $\mathcal{A} = (\Theta, m_1, ..., m_k)$ be a DS model, where $\Theta$ is a frame of discernment and the $m_i$'s are BPA's on $\Theta$; let $m = \oplus_i m_i$.*
*Then there is a formal system $\Sigma_\mathcal{A}$ and a Belief Base $\kappa_m$ on $\Sigma_\mathcal{A}$ such that, for all $A \subseteq \Theta$*
$\text{Ask}_{\Sigma_\mathcal{A}}[P_A, \kappa_m] = <x_t, x_f>$ *iff* $x_t = bel_m(A), x_f = bel_m(-A)$
*where $P_A$ is the sentence of $\mathcal{L}_{\Sigma_\mathcal{A}}$ expressing "the answer to $\mathcal{A}$ is in $A$", and $bel_m$ is the standard belief function associated to m.*

## 4. Modelling a Belief Base

We turn now to considering the problem of instantiating Belief Bases to particular cases of interest, i.e. to define specific Belief Bases for some given KR system $\Sigma$. This is made in two steps: the first step is to chose a particular way of modelling propositions, i.e. to define the elements of the set $\mathcal{P}$ and the relation $\Rightarrow$ on it; the second step consists in defining a meaning function $\|\cdot\|_\Sigma$ for the specific KR system we want to use, based on a suitable formalization of it.

A first possibility for modelling $\mathcal{P}$, fairly usual in the logical tradition, is to identify a proposition with a set of possible worlds, namely those worlds in which that proposition holds. For the sake of generality, we consider here Kripke structures[6] of the form $M = \langle S_M, D_M, V_M, \{\mathcal{R}_{M_i} \mid i \geq 0\}\rangle$, where $S_M$ is a set of states, $D_M$ is a domain of individuals, $V_M$ is a mapping from symbols of $\mathcal{L}$ and states $s \in S_M$ to elements (and sets) of $D_M$, and the $\mathcal{R}_{M_i}$'s are binary relations over $S_M$. A Kripke (or "possible") world is a pair $<M,s>$ with $s \in S_M$. $\mathcal{P}$ is then given by the power-set of the set of all Kripke worlds, equipped with the standard $\subseteq$ relation.

---

[5] The proofs of these results are given in the full paper.
[6] Other mathematical structures used to give semantics to languages could have been taken here. The reader unfamiliar with Kripke structures is referred to e.g. (Hughes & Cresswell, 1968).

Let us now consider having a KR system $\Sigma$. In order to build a meaning function $\|\cdot\|_\Sigma$ for it, we need to have a truth relation $\vDash_\Sigma$ defining the notion of truth in a world. This means that a formalization of $\Sigma$ of the form $(\mathcal{L}_\Sigma, \vDash_\Sigma)$ is needed. The $\vDash_\Sigma$ relation will in general be defined only on certain types of Kripke structures $M_\Sigma$; correspondingly, only certain Kripke worlds will be considered. We denote by $\mathcal{W}_\Sigma$ the set of worlds on which the semantics of $\Sigma$ is defined. We then let the proposition connoted by $\alpha$ in $\Sigma$ be the set of all the possible worlds in $\mathcal{W}_\Sigma$ where $\alpha$ holds according to $\vDash_\Sigma$. We define the function $\|\cdot\|_\Sigma: \mathcal{L}_\Sigma \to \mathcal{P}$ accordingly:

**Def. 4.1.** *Let $\Sigma$ and $\mathcal{W}_\Sigma$ be as above. Then, for each sentence $\alpha \in \mathcal{L}_\Sigma$, we let:* $\|\alpha\|_\Sigma =_{df} \{w \in \mathcal{W}_\Sigma \mid w \vDash_\Sigma \alpha\}$.

We can now interpret definition 3.3 in terms of possible worlds. Belief Bases on $\Sigma$ are sets of pairs $<P, m_P>$ with $P$ a subset of $\mathcal{W}_\Sigma$ and $m_P \in [0,1]$: intuitively, they represent BPA's on the set of all the possible worlds for $\Sigma$. The condition "$P \Rightarrow Q$" in the definition of $\text{Bel}_\kappa$ reads "$P \subseteq Q$" in the chosen model for $\mathcal{P}$. This condition corresponds to logical entailment between sentences of $\mathcal{L}_\Sigma$: given two sentences $\alpha$ and $\beta$, $\|\beta\|_\Sigma \subseteq \|\alpha\|_\Sigma$ is true whenever $\{w \mid w \vDash_\Sigma \beta\} \subseteq \{w \mid w \vDash_\Sigma \alpha\}$, that is whenever $\beta \vDash_\Sigma \alpha$. $\text{Ask}_\Sigma[\alpha, \kappa]$ measures the total belief committed to the set of possible worlds $\|\alpha\|_\Sigma$ (and to its complement) by the BPA $\kappa$. $\text{Tell}_\Sigma[\alpha, <x_t, x_f>, \kappa]$ adds to $\kappa$ all the intersections of the propositions already in $\kappa$ with $\|\alpha\|_\Sigma$ and with $^c\|\alpha\|_\Sigma$. Very roughly, each set of worlds $P$ in $\kappa$ is split into three subsets: those worlds in $P$ where $\alpha$ holds, those where $\alpha$ does not hold, and $P$ itself. Mass of belief is then redistributed over all the resulting sets according to Dempster's combination. Finally, $\text{Empty}_\Sigma[]$ allocates all the mass of belief to the whole set $\mathcal{W}_\Sigma$; as all and only the valid sentences of $\Sigma$ hold in all possible worlds, then all and only the valid sentences of $\Sigma$ will be believed (with unitary belief) in the empty Belief Base.

**Example 1.** We can use Belief Bases to link together DS theory and First Order Logic (FOL). Consider $(\mathcal{L}_{FOL}, \vDash_{FOL})$, where $\mathcal{L}_{FOL}$ and $\vDash_{FOL}$ are the language and the truth relation for FOL, respectively. $\mathcal{W}_{FOL}$ is the set of worlds $<M_{FOL}, s_0>$, with $M_{FOL} = \langle\{s_0\}, D, V\rangle$ such that $\langle D, V\rangle$ is a standard FOL interpretation structure. We therefore have: $\|\alpha\|_{FOL} =$

{<M,s₀> | <M,s₀>⊨_FOL α}; i.e. ||α||_FOL is the set of all first order models for α. Ask_FOL, Tell_FOL and Empty_FOL are plainly defined according to definition 3.3, by substituting ||α||_FOL to ||α||_Σ and W_FOL to U. In the resulting Belief Bases, knowledge will be represented in first order logic, and belief about it will obey DS theory. The picture below illustrates the Belief Base κ obtained by (κ₀ stands for Empty_FOL[]):

κ ≡ Tell_FOL[ dog(Alex), <0.7, 0.1>,
  Tell_FOL[ ∀x.(dog(x) ⊃ animal(x)), <0.9, 0>, κ₀]]

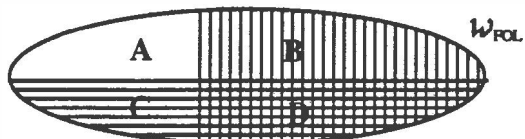

The vertically shaded part comprises worlds in which "∀x.(dog(x) ⊃ animal(x))" is true; the horizontally shaded one, worlds in which "dog(Alex)" is true; D corresponds to ||∀x.(dog(x) ⊃ animal(x)) ∧ dog(Alex)||_FOL. The first Tell allocates a mass of 0.9 to the subset B∪D, and 0.1 to the whole space A∪B∪C∪D. The second Tell combines this BPA with one in which a mass of 0.7 is allocated to C∪D, 0.1 to A∪B, and 0.2 to A∪B∪C∪D. In the resulting Belief Base κ, a mass of 0.02 is being allocated to A∪B∪C∪D, 0.01 to A∪B, 0.07 to C∪D, 0.18 to B∪D, 0.09 to B, and 0.63 to D. By definition of FOL interpretation, the sentence "animal(Alex)" will uniformly hold in D (while it will in general hold in some of—but not all—the worlds in A, B and C); this means that the only proposition in κ which entails ||animal(Alex)||_FOL is D, while no proposition entails ||¬animal(Alex)||_FOL. Hence:

Ask_FOL[animal(Alex), κ] = <0.63, 0>

**Example 2.** We continue Example 1 by showing how the Tweety problem may be formulated with Belief Bases on FOL (κ₀ is as above; we omit for short the subscript FOL):

κ₁ ≡ Tell[(∀x.(bird(x) ∧ ~excp(x)) ⊃ flier(x)), <1,0>, κ₀]
κ₂ ≡ Tell[(∀x. ~excp(x)), <0.8, 0.2>, κ₁]
κ₃ ≡ Tell[(∀x.(penguin(x) ⊃ (bird(x)∧~flier(x)))),<1,0>,κ₂]
κ₄ ≡ Tell[ bird(Tweety), <1, 0>, κ₃]
κ₅ ≡ Tell[ bird(Cippy), <1, 0>, κ₄]
κ₆ ≡ Tell[ penguin(Tweety), <1, 0>, κ₅]

Then we have, in κ₅:
  Ask[ flier(Tweety), κ₅] = <0.8, 0>
  Ask[ flier(Cippy), κ₅] = <0.8, 0>
and, in κ₆:
  Ask[ flier(Tweety), κ₆] = <0, 1>
  Ask[ flier(Cippy), κ₆] = <0.8, 0>
  Ask[ excp(Tweety), κ₆] = <1, 0>
  Ask[ excp(Cippy), κ₆] = <0, 0.8>

**Example 3.** We consider M-KRYPTON, an extension of the KRYPTON hybrid KR system (Brachman et al., 1985), developed to model the interaction between multiple agents; in particular, M-KRYPTON can express sentences like $(B_i \alpha)$, read "agent $i$ believes[7] that α"; a full description of M-KRYPTON is given in (Saffiotti & Sebastiani, 1988). Apart from its expressive power, the interest in using M-KRYPTON here lies in its semantics being defined in a possible world framework, in terms of a truth relation ⊨_MK. Belief Bases on M-KRYPTON = ($\mathcal{L}_{MK}$, ⊨_MK) are plainly given by def. 3.3, by letting ||α||_Σ = {ω∈W_MK, | ω⊨_MK α } and U = W_MK (where W_MK is the set of Kripke worlds on which ⊨_MK is defined). We illustrate the behaviour of the resulting Belief Bases on the following interesting problem:

*"After a long discussion, Robert and Alex came to a good agreement about which individuals are (or are not) human; also, most probably Alex knows that pets are animals owned by humans. Given that Alex knows that Tweety is a pet, and that Philippe owns it, will Robert believe Philippe a human?"*

κ₁≡Tell_MK[(FORALL x (B_al(Human x))↔(B_bob(Human x))), <0.8, 0>, Empty_MK[]]

κ₂≡Tell_MK[(B_al(IS Pet (VRGeneric Animal owner Human))), <0.98, 0>, κ₁]

κ₃≡Tell_MK[(B_al(AND (Pet tweety) (owner tweety philippe))), <1, 0>, κ₂]

Then: Ask_MK[(B_bob(Human philippe)), κ₃] = <0.784, 0>
Notice that the kind of knowledge represented in this example is to no respect a trivial one: the first Tell expresses a contingent relation between the beliefs held by two different agents; the second one expresses the belief of an agent toward a terminological definition; and the third one expresses the belief of an agent toward some assertions. All these types of knowledge are captured by the mechanisms of M-KRYPTON. By contrast, a formalization of this problem in a standard DS formalism appears far from being obvious.

We switch now to a different way of modelling propositions. There are two reasons for doing this: to illustrate an alternative way to model Belief Sets (needed, e.g., if a proof-theoretic account of the KR system we want to use is available, rather than a model-theoretic one); and to get closer to the practical side of Belief Bases (as we will discuss in the next Section). We will model propositions in terms of more "tangible" syntactical structures, rather than semantical ones. Informally, given a sentence α of $\mathcal{L}_\Sigma$, we consider those sets of sentences of $\mathcal{L}_\Sigma$ such that α is deducible from them in Σ; the proposition connoted by α in Σ consists of the collection of all these sets of sentences.

More formally, given a proof-theoretic account of a KR system Σ = ($\mathcal{L}_\Sigma$, ⊢_Σ), we define a *possible argument* in Σ to be any consistent set of sentences of $\mathcal{L}_\Sigma$. In order to avoid unnecessary complexities, we also re-

---

[7] The traditional epistemic notion of (categorical) belief is meant here, with no reference to Dempster-Shafer's belief.

quire possible arguments to be non-redundant, i.e. none of the sentences β in a possible argument π must be provable by the other sentences in π through $\vdash_\Sigma$. We denote by $Q_\Sigma$ the set of all possible arguments in Σ. We then redefine the function $\|\cdot\|_\Sigma: \mathcal{L}_\Sigma \to \mathcal{P}$ in terms of of possible arguments[8]:

**Def.4.2.** *Let Σ and $Q_\Sigma$ be as above. Then, for each sentence α of $\mathcal{L}_\Sigma$, we let $\|\alpha\|_\Sigma =_{df} \{\pi \in Q_\Sigma \mid \pi\vdash_\Sigma \alpha\}$.*

The algebra of propositions $\mathcal{P}$ has now sets of subsets of $\mathcal{L}_\Sigma$ as its elements, and again ⊆ as its partial order. Hence, definition 3.3 may now be read in the following way. Belief Bases represent BPA's on $Q_\Sigma$. The condition "$P \Rightarrow Q$" ("$P \subseteq Q$") in the definition of $Bel_\kappa$ should now be interpreted in terms of deducibility: given two sentences α and β of $\mathcal{L}_\Sigma$, $\|\beta\|_\Sigma \subseteq \|\alpha\|_\Sigma$ is true whenever $\{\pi \mid \pi\vdash_\Sigma \beta\} \subseteq \{\pi \mid \pi\vdash_\Sigma \alpha\}$; this means that, whatever argument is valid among the possibilities in $\|\beta\|_\Sigma$ (in particular, {β} itself), it must prove α as well, that is $\beta\vdash_\Sigma \alpha$. $Ask_\Sigma[\alpha,\kappa]$ measures the total belief committed to $\|\alpha\|_\Sigma$ (and to its complement) by the BPA κ. $Tell_\Sigma[\alpha,<x_t,x_f>,\kappa]$ considers all the sets of possible arguments in κ; each of them is intersected with the sets in $\|\alpha\|_\Sigma$ and in $^c\|\alpha\|_\Sigma$ in the course of Dempster's combination. Very roughly, each set P in κ is split into three subsets: those possible arguments in P that prove α, those that prove ~α, and P itself. Mass of belief is then redistributed over all the resulting sets of possible arguments. Finally, $Empty_\Sigma[]$ allocates all the mass of belief to the whole $Q_\Sigma$; as $\pi\vdash_\Sigma \alpha$ holds for each π if and only if α is a theorem of Σ, then all and only the theorems of Σ will be believed (with unitary belief) in the empty Belief Base.

**Example 4.** Any KR system for which a proof theory is provided can be used to define Belief Bases, by defining $\|\cdot\|_\Sigma$ as in def. 4.2. So, we could textually repeat here Example 2 by using $\vdash_{FOL}$ instead of $\models_{FOL}$: the behaviour of the resulting Belief Bases would be exactly the same. As a matter of fact, the particular way propositions are modelled (and correspondingly the way—model theoretic or proof theoretic—in which the formalization of the KR system is given) is completely transparent to the user of the resulting Belief Bases.

## 5. A Stairway to Concreteness

We focus in this section on the practical side of Belief Bases: we discuss a possible architecture for a modular system embodying Belief Bases, and we consider a possible approach to its implementation. The proof-theoretic modelling of Belief Bases discussed above provides a hook for pursuing our program. In fact, switching from possible worlds to possible arguments allows us, by going down to the syntactic level, to get closer to the practical side of our Belief Bases[9]. We sketch a first architecture for our system:

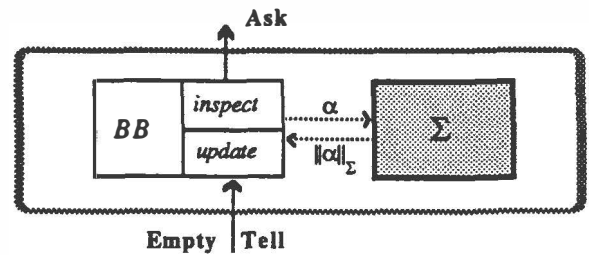

In this architecture, the Belief Base is considered as an internally stored data structure (BB); as a consequence, we use primitive operations $Empty_\Sigma[]$, $Ask_\Sigma[\alpha]$ and $Tell_\Sigma[\alpha,<x_t,x_f>]$ where the Belief Base argument is omitted (reference to the internal BB being implicit). The *inspect* and *update* modules implement these operations, i.e. they implement the interface mechanism and the DS calculus on the propositions in BB. The KR system Σ constitutes a neatly distinct module (possibly an already existing system) whose role is to provide $\|\alpha\|_\Sigma$ for each sentence α of the KR language. When $\|\alpha\|_\Sigma$ is a set of possible arguments for α, this role is not completely detached from a plausible behaviour for a KR system; yet, as this set includes <u>all</u> the possible arguments according to the logic Σ, irrespective of what is actually believed in BB, $\|\alpha\|_\Sigma$ is in general a computationally intractable object[10]. Two steps may be undertaken at this stage:

1. to only consider possible arguments which are "reasonable" with respect to what is actually believed in the Belief Base; and
2. to let Σ provide "fragments" of possible arguments, generated while performing inferences; the reconstruction of full possible arguments from these fragments is then performed outside Σ.

---

[8] When $\mathcal{L}_\Sigma$ comprises a conjunction operator ∧ satisfying "$(\alpha,\beta)\vdash\gamma$ iff $(\alpha\wedge\beta)\vdash\gamma$", and a disjunction operator ∨ satisfying "$(\alpha)\vdash\gamma$ or $(\beta)\vdash\gamma$ iff $(\alpha\vee\beta)\vdash\gamma$", a (finite) set of possible arguments $\{\pi_i \mid \pi_i=(\beta_{i1},...,\beta_{iN_i})\}$ is equivalent to the sentence of $\mathcal{L}_\Sigma$ "$\vee_i(\wedge_j \beta_{ij})$". In this case, we could represent propositions by sentences of $\mathcal{L}_\Sigma$. However, this is not true in general.

[9] Formally, this change is pretty irrelevant: possible worlds actually correspond to classes of equivalence of elements of $Q_\Sigma$.

[10] Notice that the full set $Q_\Sigma$ for a language with n atomic sentences will in general comprise at least $2^{2^n}$ elements.

Step 1 implies that $\Sigma$ must access BB to decide which (possible) inferences to draw in order to find "reasonable" possible arguments (alternatively, knowledge in BB may be cached in a Knowledge Base internal to $\Sigma$). To exemplify, only those possible arguments in $Q^*$ could be considered at any moment, where $Q^*$ is the subset of $Q_\Sigma$ comprising only the sentences which have occurred up to now in a Tell operation. The effect of this is to drastically reduce the average number and size of possible arguments.

Step 2 greatly weakens the demands on $\Sigma$, switching to a requirement more easily fulfilled by typical KR systems. In a plausible scenario, fragments could correspond to reports of single inference steps: hence, they might be seen as ATMS (deKleer, 1986) justifications, and the burden of reconstructing full possible arguments from them might be carried on by an ATMS linked to $\Sigma$. Intuitively, and very roughly, the possible arguments for $\alpha$ correspond to the environments in the label built by the ATMS for the node[11] $\gamma_\alpha$, given a set of justifications communicated by $\Sigma$ while looking for possible ways of deducing $\alpha$. Asking $\alpha$ means computing a belief value for the labels of $\gamma_\alpha$ and of $\gamma_{\neg\alpha}$, by summing up the mass given to each environment in them. Telling $\alpha$ means justifying $\alpha$ and $\neg\alpha$ by a pair of assumptions A and Ã with the appropriate mass values attached. This algorithm is summarized below:

---

**Tell$_\Sigma[\alpha, <x_t, x_f>]$**

1. Ask $\Sigma$ to try to deduce both $\alpha$ and $\neg\alpha^{(i)}$; in the deduction, each inference $\beta_1, ..., \beta_{k-1} / \beta_k$ performed by $\Sigma$ will be communicated to the ATMS as a justification $\gamma_{\beta_1}, ..., \gamma_{\beta_{k-1}} \Rightarrow \gamma_{\beta_k}$.
2. Create new ATMS assumptions $\Gamma_A$ and $\Gamma_{\tilde{A}}$, and communicate justifications $\Gamma_A \Rightarrow \gamma_\alpha$; $\Gamma_{\tilde{A}} \Rightarrow \gamma_{\neg\alpha}$; and $\Gamma_A, \Gamma_{\tilde{A}} \Rightarrow \bot$.
3. Store $x_t$ and $x_f$ as the masses of assumptions $\Gamma_A$ and $\Gamma_{\tilde{A}}$.

**Ask$_\Sigma[\alpha]$**

1. The same as step 1 of Tell.
2. Ask the ATMS the labels of $\gamma_{\neg\alpha}$ and of $\gamma_\alpha$
3. Compute the value of Bel($\alpha$) by summing up the masses of all the environments in the label of $\gamma_\alpha$ (products of the masses of all the assumptions comprising it); do the same for Bel($\neg\alpha$)$^{(ii)}$.

*Notes:*
i) "try to deduce" *may involve finding all the potential deductions, or only those grounded on believed knowledge*
ii) *Some precaution must be taken in order for the environments to be disjoint (cf. Laskey-Lehner, 1989).*

---

Notice that the number of computations performed by the algorithm above is, in general, large. Usual ATMS algorithms for computing labels are exponential in the number of assumptions; also, the number of elements in the labels tend to grow exponentially with the number of assumptions. As the number of assumptions in the algorithm above is proportional to the number of Tell$_\Sigma$ operations, the complexity of computing Ask$_\Sigma[\alpha]$ will in general be exponential in the number of Tell$_\Sigma$ performed up to now[12]. This should not surprise us, as the computational complexity of the obvious algorithm for computing Dempster's combination is exponential in the size of the frame of discernment.

## 6. Discussion and Related Work

Workers on DS theory have traditionally been rather unconcerned with the linguistic structures used to represent knowledge. The usual attitude consists in leaving to the user the burden of expressing the knowledge relevant to her problem in the mathematical framework provided by DS theory, i.e. by means far away from the languages most commonly used in knowledge representation. This appears to be a general attitude in the field of uncertain reasoning. In some case, some attention has been devoted to the linguistic aspect, as in (Zadeh, 1989); however, Zadeh advocates the use of Fuzzy Logic <u>as</u> a KR tool, while we suggest to combine DS theory <u>with</u> a KR tool. Other frameworks (like some of the existing Expert System Shells) have occasionally been developed in which both the need of representing knowledge and that of representing uncertainty have been taken into considerations. In them, however, uncertainty is either part of the language, or it is attached to sentences of the language. In our approach, on the contrary, uncertainty is attached to the knowledge itself, rather than to the linguistic structures used to represent it. Notice that this approach carries with it an interpretation of uncertainty (or, more precisely, *epistemic* uncertainty) as meta-knowledge about the validity of our knowledge with respect to an intrinsically certain reality: let us consider the two statements

a) I am 80% sure that birds fly
b) 80% of birds fly

statement (a) illustrates our notion of uncertain knowledge: we can see it as a categorical implication (e.g. "$\forall x.\text{bird}(x) \rightarrow \text{fly}(x)$") accompanied by (epistemic) uncertainty information; on the contrary, we regard (b) as an example of categorical knowledge which refers to a statistical (but epistemically sure) fact.

The main outcome of this paper is the definition of Belief Bases, abstract data types capable of representing Dempster-Shafer's belief about structured knowledge expressed in some KR language. The "good behaviour" of Belief Bases, as shown by lemma 3.1 and theorem 3.1, guarantees that the properties of both the KR system used and of DS theory are preserved in the combination. From the point of view of workers in knowl-

---

[11] We write $\gamma_\alpha$ to refer to the ATMS node associated to the value "true" for $\alpha$; correspondingly, we also associate an ATMS node $\gamma_{\neg\alpha}$ to the value "false" for $\alpha$. $\Gamma_A$ denote as usual the ATMS node associated with the (truth of the) assumption A.

[12] We would obtain the same result by considering the general proof-theoretic definition, with the above restriction on the $Q^*$.

edge representation, Belief Bases constitute a tool for attaching a well-grounded and powerful treatment of uncertainty to a KR system. From the point of view of people fond of DS theory, they provide a means for extending the applicability of DS theory to kinds of knowledge and problems whose formalization in a DS framework would otherwise be far from obvious. To be sure, the addition of simply the full first order logic to DS theory (cf. Example 1) is already a non-trivial result in terms of the expressiveness of the theory. For instance, the following situation would be hardly expressible in a standard use of DS theory (e.g. the multivariate approach):

$\kappa_1 = \text{Tell}_{FOL}[\forall x.(\exists y.\text{spouse}(x,y) \rightarrow \text{married}(x)), <0.98, 0>, \text{Empty}_{FOL}[]]$

$\kappa_2 = \text{Tell}_{FOL}[\text{spouse}(\text{Robert}, \text{Alice}), <0.7, 0>, \kappa_1]$

$\text{Ask}_{FOL}[\text{married}(\text{Robert}), \kappa_2] = <0.686, 0>$

Moreover, using KR languages other than FOL allows us to express belief about more complex notions (e.g. epistemic attitudes of multiple agents—cf. Example 3) or finer distinctions (e.g. the distinction between *terminological* and *assertional* knowledge—same example).

The possible world formalization given in this work is related to other possible world accounts given to Dempster-Shafer's theory (e.g. Ruspini, 1986; Fagin and Halpern, 1989). However, our focus here is the linking of DS theory to an arbitrary KR system, by using propositions as a formal bridge: possible worlds are just one possible choice for modelling propositions. Moreover, the above referred accounts are normally restricted to considering the propositional case, while the approach presented here allows us to extend DS theory to any suitable formal language. On a different side, the ATMS-based algorithm proposed for Belief Bases is extremely similar to some recent proposal to use ATMS for implementing DS theory (D'ambrosio, 1988; Laskey & Lehner, 1989). This similarity is intriguing, as both the starting point and the goal of the authors above are apparently different from the ones here: in their proposals, a DS model is mapped into a set of ATMS justifications plus a set of ATMS assumptions with attached probability values. The ATMS is used as a mechanism to perform belief propagation in a symbolic way; computing then the probability of the ATMS label for the ATMS node representing a value for a variables gives us the belief in that variable taking that value. On the contrary, we are not concerned at all with fitting a general DS model into an ATMS: rather, ATMS is coupled in a standard way to a KR system (or Problem Solver) and used as a computational mechanism to reconstruct the "possible arguments" (ATMS environments) for a node given the set of justifications produced by the KR system in the deduction process. These possible arguments constitute, independently of the way they are computed, the basis for the computation of belief according to definition 3.3.

The approach presented here is extensible to other uncertainty calculi. In particular, it is possible to define a formal framework in which both a specific KR system and a specific uncertainty calculus are "plugged in" as modules. This general framework is presented in (Saffiotti, 1990a), while (Saffiotti, 1990b) elaborates on its architecture on the lines hinted at in Section 5.

**Acknowledgements.** The present research has benefit from discussions with (and comments from) Fabrizio Sebastiani, Yen-Teh Hsia, Robert Kennes, Bruno Marchal, Philippe Smets and Nic Wilson.

## References

Brachman, R.J. and Levesque, H.J. (1982) "Competence in Knowledge Representation", Proc. of AAAI-82.

Brachman, R.J., Pigman Gilbert, V. and Levesque, H.J. (1985) "An Essential Hybrid Reasoning System: Knowledge and Symbol Level Accounts of Krypton", Proc. of IJCAI-85: 532-539.

D'Ambrosio, B. (1988) "A Hybrid Approach to reasoning Under Uncertainty", *International Journal of Approximate Reasoning* 2:29-45.

de Kleer, J. (1986) "An Assumption-Based Truth Maintenance System" *Artificial Intelligence* 28: 127-162.

Fagin, R. and Halpern, J.Y. (1989) "Uncertainty, Belief and Probability", Proc. of IJCAI-89: 1161-1167.

Hughes, G. E. and Cresswell, M.J. (1968) *An introduction to modal logic* (Methuen & Co., UK).

Israel, D.J. and Brachman, R.J. (1981) "Distinctions and Confusions: a Catalogue Raisonné", Proc. of IJCAI-81.

Kong, A. (1986) "Multivariate Belief Functions and Graphical Models", PhD Thesis (Harvard Univ., CA).

Laskey, K.B. and Lehner, P.E. (1989) "Assumptions, Beliefs and Probabilities", *Artificial Intelligence* 41(1).

Levesque, H.J. (1984) "Foundations of a Functional Approach to Knowledge Representation", *Artificial Intelligence* 23: 155-212.

Ruspini, E.H. (1986) "The Logical Foundations of Evidential Reasoning", TN 408, SRI International.

Saffiotti, A. (1990a) "A Hybrid Framework for Representing Uncertain Knowledge", to appear in: Procs of the Eighth AAAI Conf.(Boston, MA).

Saffiotti, A. (1990b) "A Hybrid Belief System for Doubtful Agents", Proc. of the Third IPMU Conf. (Paris, France).

Saffiotti, A. and Sebastiani, F. (1988) "Dialogue Modelling in M-KRYPTON, a Hybrid Language for Multiple Believers", Proc. of the IEEE Conf. on AI Applications.

Shafer G. (1976) *A Mathematical Theory of Evidence* (Princeton University Press, Princeton).

Smets, P. (1988) "Belief Functions", in: Smets P., Mamdani E.H., Dubois D. and Prade H. (Eds.) *Non-Standard Logics for Automated Reasoning* (Academic Press, London).

Zadeh, L.A. (1989) "Knowledge Representation in Fuzzy Logic", *IEEE Trans. on Knowledge and Data Engineering* 1(1):89-100.

# Session 8:

## Second Poster Session

A Hierarchical Approach to Designing Approximate Reasoning-
Based Controllers for Dynamic Physical Systems
*H. Berenji, Y.Y Chen, C.C. Lee, J.S. Jang and S. Murugesan*

Evidence Combination and Reasoning and Its
Application to Real-World Problem-Solving
*L.W. Chang and R.L. Kashyap*

On Some Equivalence Relations between Incidence
Calculus and Dempster-Shafer Theory of Evidence
*F. Correa da Silva and A. Bundy*

Using Belief Functions for Uncertainty Management and
Knowledge Acquisition: An Expert System Application
*M. Deutsch-McLeish, P. Yao and T. Stirtzinger*

An Architecture for Probabilistic Concept-Based
Information Retrieval
*R.M. Fung, S.L. Crawford, L.A. Appelbaum and R.M. Tong*

Amplitude-Based Approach to Evidence Accumulation
*A.J. Hanson*

A Probabilistic Reasoning Environment
*K.B. Laskey*

On Non-monotonic Conditional Reasoning
*H.T. Nguyen*

Decisions with Limited Observations
over a Finite Product Space: the Klir Effect
*M. Pittarelli*